\documentclass[runningheads]{llncs}
\usepackage{graphicx}
\usepackage{comment}
\usepackage{amsmath,amssymb} % define this before the line numbering.
\usepackage{color}

\usepackage{mathtools}
\usepackage{makecell}
\usepackage{multirow}
\usepackage{xcolor}
\usepackage{colortbl}
\usepackage{boldline}
\usepackage{algorithm}
\usepackage{algorithmicx}
\usepackage{algcompatible}
\usepackage{algpseudocode}
\usepackage{caption}
\begin{document}
\pagestyle{headings}
\mainmatter
\def\ECCVSubNumber{}

\title{To Balance or Not to Balance: \\A Simple-yet-Effective Approach for Learning with Long-Tailed Distributions}

%******************

\titlerunning{A Simple-yet-Effective Approach for Learning with LT Distributions.}

\author{Junjie Zhang\inst{1} \and
Lingqiao Liu\inst{1} \and
Peng Wang\inst{2} \and
Chunhua Shen\inst{1}
\institute{The University of Adelaide, Australia \and
University of Wollongong, Australia}
\email{firstname.lastname@adelaide.edu.au},
\email{pengw@uow.edu.au}
}
\authorrunning{J. Zhang, L. Liu, P. Wang \& C. Shen.}

%******************
\maketitle

\begin{abstract}
Real-world visual data often exhibits a long-tailed distribution, where some ``head'' classes have a large number of samples, yet only a few samples are available for ``tail'' classes. Such imbalanced distribution causes a great challenge for learning a deep neural network,  which can be boiled down into a dilemma: on the one hand, we prefer to increase the exposure of tail class samples to avoid the excessive dominance of head classes in the classifier training. On the other hand, oversampling tail classes makes the network prone to over-fitting, since head class samples are often consequently under-represented. To resolve this dilemma, in this paper, we propose a simple-yet-effective auxiliary learning approach. The key idea is to split a network into a classifier part and a feature extractor part, and then employ different training strategies for each part. Specifically, to promote the awareness of tail-classes, a class-balanced sampling scheme is utilised for training both the classifier and the feature extractor. For the feature extractor, we also introduce an auxiliary training task, which is to train a classifier under the regular random sampling scheme. In this way, the feature extractor is jointly trained from both sampling strategies and thus can take advantage of all training data and avoid the over-fitting issue. Apart from this basic auxiliary task, we further explore the benefit of using self-supervised learning as the auxiliary task. Without using any bells and whistles, our model achieves superior performance over the state-of-the-art solutions.
\end{abstract}

\section{Introduction}
\label{intro}

For many real-world visual datasets, visual concepts often occur with a long-tailed distribution, that is, some ``head'' classes have abundant examples, while only a few samples are available for ``tail'' classes \cite{liu2019large,khan2019striking,cui2019class}. Such an imbalanced data distribution poses a great challenge for training a deep neural network since standard stochastic gradient descent (SGD) \cite{bottou2010large} tends to make the network ignore the tail classes due to the chance of sampling from a given tail class can be much lower compared to samples from head classes.

A straightforward solution to the above issue seems to increase the chance of sampling from tail classes by balancing the sampling ratio per class. For example, one can force data within a mini-batch to be evenly sampled from each class. However, this naive solution may lead to a side-effect of making head classes under-represented and increase the risk of over-fitting a deep neural network. Therefore, there is a dilemma in deciding whether or not to balance the training samples from each class. 
Existing approaches tackle this dilemma by carefully designing re-sampling strategies \cite{haixiang2017learning,chawla2002smote,tahir2009multiple}, class-dependent cost functions \cite{huang2016learning,mahajan2018exploring}, or feature regularisation approaches \cite{zhang2017range,liu2019large} \textit{etc.} In this work, we propose a simple auxiliary learning approach without any bells and whistles, yet surprisingly, this simple approach achieves the state-of-the-art performance.  

\begin{figure}[t]
    \centering
    \includegraphics[width=1.02\linewidth]{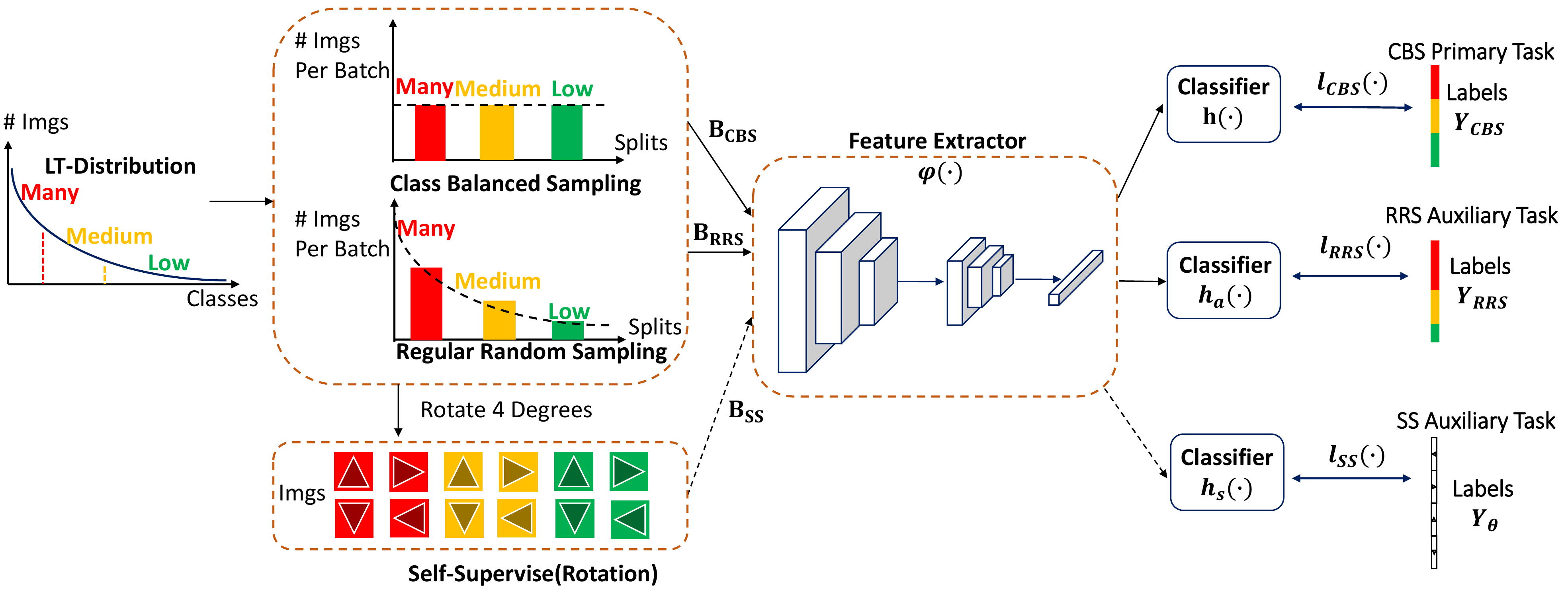}
    \caption{The overview of the proposed framework. The class-balanced sampler is used in the primary classification task CBS, where balanced input batches $B_{\text{CBS}}$ is fed to the CNN feature extractor $\varphi(\cdot)$ and classifier $h(\cdot)$. The regular random sampling auxiliary task RRS is utilised to prevent the ill-fitting of representation learning. Moreover, rotation based self-supervised learning as an additional auxiliary task is investigated to further improve the generalisation of the learned features. Since all auxiliary tasks are proposed to adjust the representation learning, they share the feature extractor $\varphi(\cdot)$ with the primary task, and separate classifiers are trained. The $\varphi(\cdot)$ and classifier $h(\cdot)$ from the primary task are retained for prediction.}
    \setlength{\belowcaptionskip}{-10pt}
    \label{3_framework}
\end{figure}

Our solution is based on the fact that a deep neural network can be decomposed into a feature extractor part and a classifier part, and these two parts can be trained with different strategies. Specifically, to prevent the training process from being dominated by head classes, we train the whole network with the class-balanced sampling (CBS) strategy. To avoid the risk of over-fitting, for the feature extractor part, we also create an auxiliary training task, that is, to learn a classifier under the regular random sampling (RRS) scheme. In this way, the feature extractor part, which consists of most parameters of a deep network, is trained with both sampling strategies. Therefore, it can take full advantage of all the training data. In addition to using different sampling strategies to create the auxiliary task, we also explore to use self-supervised learning as an additional auxiliary task to further enhance the representation learning, and we show that this can lead to promising results.
In summary, the main contributions of our method are as follows:
\begin{itemize}

\item We propose a simple-yet-effective auxiliary learning approach to address the dilemma of balancing the head and tail classes for long-tailed visual recognition. 

\item We propose to utilise self-supervised learning as an additional auxiliary task to improve the generalisation of image features.

\item We conduct comprehensive experiments on two long-tailed datasets to evaluate the effectiveness of the proposed approach. The experimental results demonstrate that our method outperforms state-of-the-art solutions. 
\end{itemize}
\section{Related Works}
\subsection{Imbalanced Classification}
We roughly group the related works of imbalanced image classification into four categories, including the re-sampling of training data, learning discriminative features, cost-sensitive learning and transfer learning.

\textbf{Re-Sampling. }
Re-sampling aims at alleviating the negative effect of skewed distribution by artificially balancing the samples among all classes. In general, there are two types of commonly used strategies, namely over-sampling and under-sampling \cite{haixiang2017learning}. 
Over-sampling focuses on augmenting the minority class either by duplicating the samples or generating synthetic data via interpolation \cite{chawla2002smote}. On the other hand, the under-sampling scheme achieves data balance by discarding part of majority classes~\cite{tahir2009multiple}. In \cite{wang2019dynamic}, a curriculum learning approach is proposed to dynamically adjust the sampling from imbalance to balance with hard example mining.

\textbf{Learning Discriminative Features.}
Instead of reshaping the data distribution, some methods tackle the imbalanced image classification by generating discriminative features. Metric learning approaches, including pair-wise contrastive loss \cite{sun2014deep}, triplet loss \cite{schroff2015facenet} and quintuplet loss \cite{huang2016learning} \textit{etc.} accompanied by hard-mining, can be used to explore the sample relationships within the input batch. Range loss \cite{zhang2017range} and centric loss \cite{wen2016discriminative} learn discriminative feature space by constraining class prototypes. In \cite{hayatgaussian}, authors propose a max-margin loss to consider both classification and clustering performances. Most recently, Liu \textit{et. al.} \cite{liu2019large} propose to utilise the memory network to train class prototypes and further enhance image features. Their method achieves state-of-the-art performances on the long-tailed open-world recognition. 

\textbf{Cost-Sensitive \& Transfer Learning.}
Cost-sensitive learning addresses the imbalanced distribution by adjusting costs for misclassifications on classes with respect to sample frequencies. Normally, the inverse of class frequencies \cite{huang2016learning} and its smoothed versions \cite{mahajan2018exploring} are used to re-weight the loss function. In \cite{khan2019striking}, authors propose to eliminate the decision boundary bias by incorporating Bayes uncertainty estimation, while in \cite{cui2019class}, the effective number of samples are calculated to construct a balanced loss, and a label-distribution-aware margin loss is proposed in \cite{cao2019learning}. Another line of works focus on transferring the knowledge \cite{ouyang2016factors,DBLP:conf/cvpr/GidarisK18} from head to tail classes through different learning algorithms, such as meta-learning \cite{wang2017learning} and unequal training \cite{zhong2019unequal} \textit{etc}.

\subsection{Auxiliary \& Self-Supervised Learning.}
Auxiliary learning is designed for assisting the primary task by simultaneously optimising relevant auxiliary tasks. It has been adopted to benefit various tasks, such as speech recognition \cite{liebel2018auxiliary}, image classification \cite{liu2019self} and depth estimation \cite{mahjourian2018unsupervised} \textit{etc.} Our work can also be regarded as an auxiliary learning approach, where an auxiliary classification task is introduced to alleviate over-fitting. 

Self-supervised learning \cite{gidaris2018unsupervised} is one type of unsupervised feature learning, which defines proxy tasks to inject self-supervision signals for representation learning. The image itself contains abundant structural information to be explored \cite{kolesnikov2019revisiting}, such as predicting low-level visual cues, or relative spatial locations of patches. In \cite{gidaris2019boosting,su2019does}, self-supervised learning is employed to assist the few-shot image classification. We have a similar motivation to explore the self-supervised learning for improving the generalisation of image features.
\section{Approach}
\label{method}

The extremely imbalanced data distribution poses a great challenge for training a deep network, which can be boiled down to a dilemma of balancing the training of head and tail classes. In this section, we first describe this dilemma that motivates our method. Then we propose our simple-yet-effective solution to this dilemma, followed by its further extension.

\subsection{A Dilemma of Balancing Head \& Tail Training}
\begin{figure}[t]
	\centering
	\resizebox{\linewidth}{!}{
		\begin{tabular}{cc}
			\includegraphics[width=0.5\linewidth]{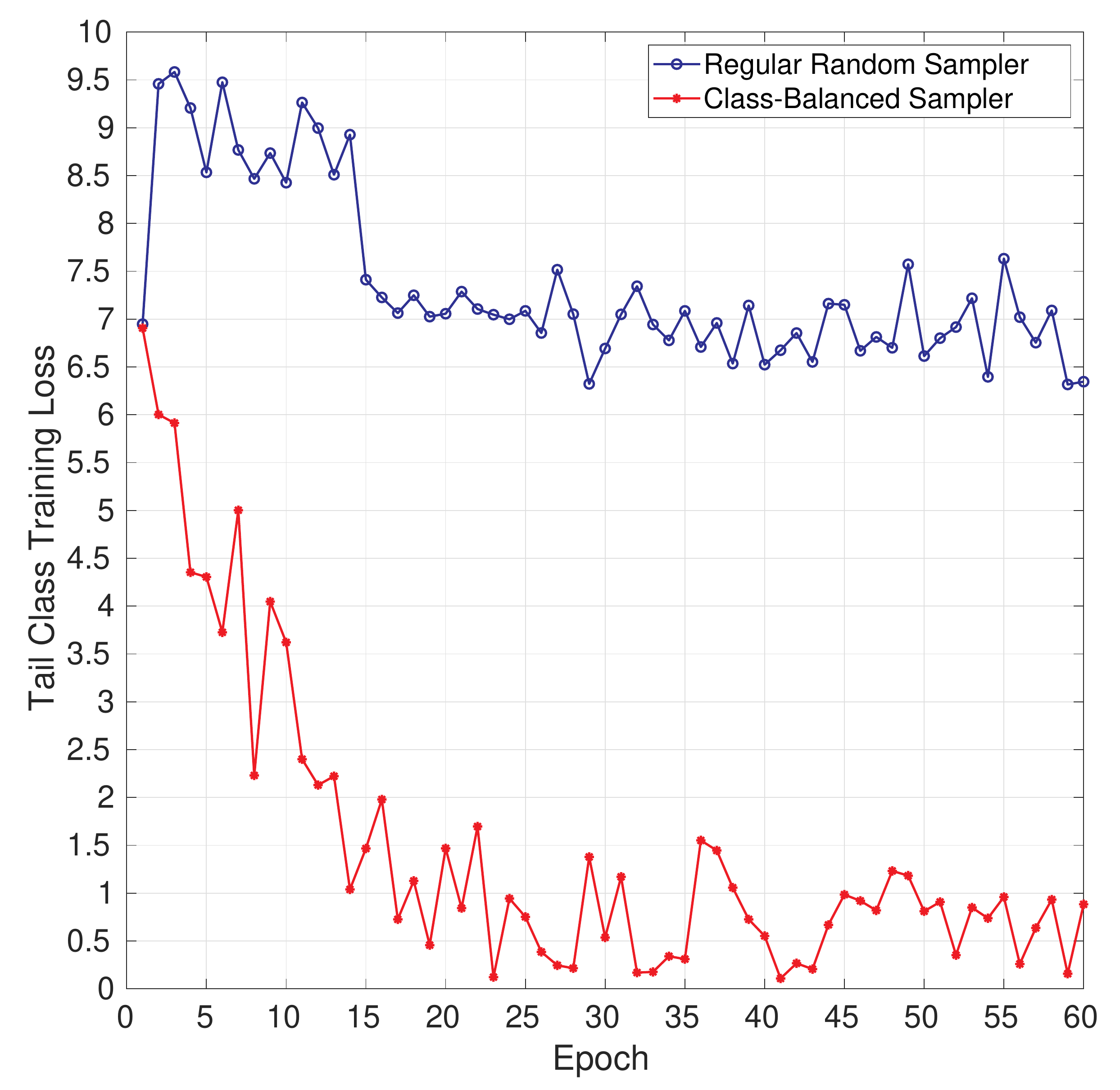} & 
			\includegraphics[width=0.5\linewidth]{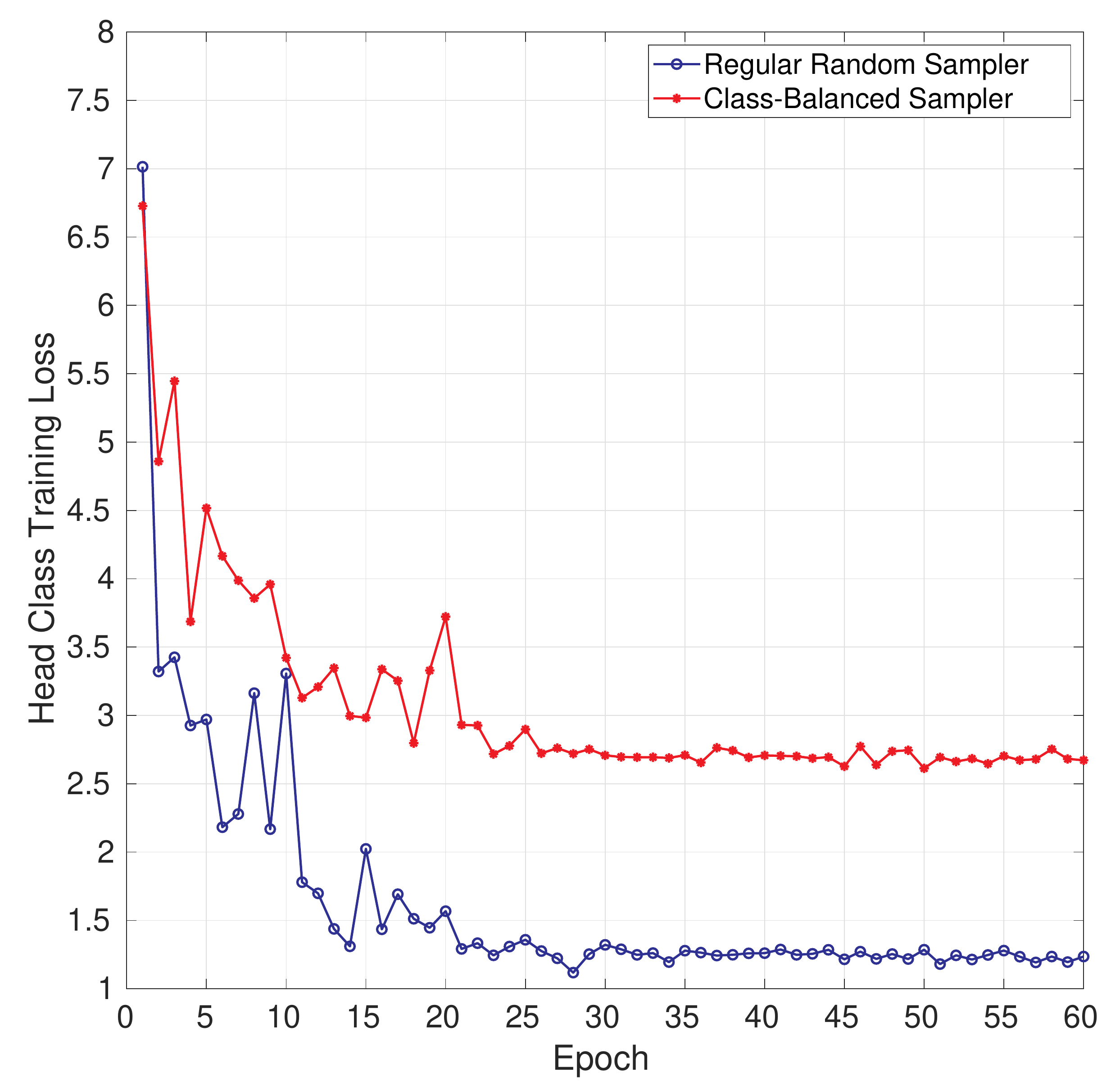}\\
	\end{tabular}}
	\caption{The mean training losses of using the regular random sampler (RRS) and the class-balanced sampler (CBS) on one tail and head class in the ImageNet-LT dataset are computed for demonstration. The left one is for the tail class, while the right one is for the head class. RRS and CBS are shown in blue and red respectively. No significant decrease is observed for RRS on the tail class, while CBS leads to a larger training loss than RRS on the head class.
	}
	\label{2_shot_trainloss}
\end{figure}

Formally, let $\mathcal{I} = \{I_i\}$ be the set of images, $\mathcal{Y} = \{y_i\}$ be the label set, where $|\mathcal{I}|=N, |\mathcal{Y}|=C$. For a multi-class classification problem, the standard training objective takes the following form:

\begin{equation}
    L = \frac{1}{N}\sum^{C}_{c=1}\sum^{N_c}_{k=1}l(h(I_{ck}), y_{ck}),
    \label{erm_class}
\end{equation}
where $l(\cdot)$ is the loss function and $h(\cdot)$ is the to-be-learned classifier. In a long-tailed training set, the number of images per class $N_c$ varies from abundant (more than thousands) to rare (a few shots). In this case, the total loss is dominated by losses from classes with many samples. In the context of deep learning, Eq. \ref{erm_class} is usually optimised by using the stochastic gradient descent (SGD) method. During each training iteration, a batch of samples, which are randomly drawn from the whole training set, are fed into the neural network. The chance of sampling a tail class sample can be meagre because of its low proportion in the training set. Consequently, the loss incurred from the tail-class samples are usually ignored during training. The blue curve in the left Fig. \ref{2_shot_trainloss} shows the training loss of one tail class under the regular random sampling. During the course of training, we cannot see a significant decrease in its value. 

A straightforward solution to this issue is to adjust the chance of sampling a tail class sample. Inspired by \cite{shen2016relay}, we can adopt a \textit{class-balanced sampling strategy} (CBS): each sampled mini-batch consists of data from $k$ classes with those classes being randomly sampled from the total list of classes, and for each sampled class, we then randomly sample the same amount ($Z$) of images.   
By using this strategy, the classifier can focus more on tail classes. The red curve in the left Fig. \ref{2_shot_trainloss} shows the impact of this strategy on the tail class. As seen, the training loss for the same tail class significantly decreases during training. 

However, class-balanced sampling strategy comes at the cost of ill-fitting. To train a deep neural network, sufficient (and diverse) samples are necessary to guarantee good generalisation performance. In class-balanced sampling, the impact of tail classes becomes more prominent. Thus, compared to the regular random sampling, tail classes have a stronger influence in learning \textit{all the parameters} in the deep neural network. This is risky since the number of samples in tail classes is too small to train so many parameters.
On the other hand, CBS over-samples tail classes. It, in effect, under-samples head classes, which makes them under-represented and further increases the risk of over-fitting the network. The red curve in the right Fig. \ref{2_shot_trainloss} shows the impact of class-balanced strategy on one head class. As seen, CBS leads to a larger training loss than that in regular random sampling. 

In summary, extremely imbalanced data distribution leads to a dilemma in training a network: increasing the influence of tail classes during training, e.g. through class-balanced sampling, is necessary since otherwise the tail classes will be ignored. However, on the other hand, it increases the risk of over-fitting a deep neural network.

\subsection{A Simple-yet-Effective Solution}\label{sect:basic solution}

Existing methods tackle this dilemma by carefully choosing sampling ratio per class \cite{haixiang2017learning,chawla2002smote,tahir2009multiple}, designing class-dependent cost functions \cite{huang2016learning,mahajan2018exploring}, or feature regularisation schemes \cite{zhang2017range,liu2019large} \textit{etc.} 

In this paper, we propose a much simpler approach to solve this dilemma. Our solution is based on the fact that a deep neural network can be decomposed into a feature extractor $\varphi(\cdot)$ and a classifier $h(\cdot)$. This decomposition allows us to adopt different training strategies for those two parts: the classifier is trained with the loss introduced in the CBS scheme; the feature extractor is trained with the loss introduced in the CBS scheme and also a regular random sampling (RRS) scheme. 

Specifically, our method is realized by constructing another classifier $h_a(\cdot)$ in addition to the original one $h(\cdot)$. Both $h(\cdot)$ and $h_a(\cdot)$ are attached to the same feature extractor $\varphi(\cdot)$ and trained jointly, as shown in Figure \ref{3_framework}. The key difference between $h(\cdot)$ and $h_a(\cdot)$ is that we train $h_a(\cdot)$ without using class-balanced sampling but regular random sampling. In such a design, the classifier is solely affected by CBS training, while the feature extractor is learned from losses of both CBS and RRS schemes. Therefore, the head classes information compromised by CBS can be recovered through the gradient back-propagated from $h_a(\cdot)$. In this sense, the feature extractor can take advantage of the information of the full dataset and the over-fitting issue can be avoided. Note that the feature extractor consists of the majority of the model parameters, and the classifier $h(\cdot)$ only involves much fewer parameters. Thus if the feature extractor does not over-fit the training data, the entire model is less likely to have the over-fitting issue.

\textbf{Remark: }\textbf{1)} Due to the auxiliary training task in $h_a(\cdot)$, head classes tend to have a stronger influence on the training of the feature extractor. One may question if this brings any side-effects as in the case of training the entire classification model with the regular random sampling strategy. We argue that the side-effects may not be severe since the feature extractor is more robust to the mismatching between the training target and deployment target, e.g., feature extractor trained on a set of classes can be reused for the classification of other sets of classes.
\textbf{2)} After training, $h_a(\cdot)$ is discarded, and only $h(\cdot)$ is used for classification. Thus our method does not use more parameters at the test time.

\subsection{Extension: Exploring A Better Feature Representation}
The key idea of our approach is to use an auxiliary training task to learn a robust feature representation, which overcomes the side-effect of CBS. In the above discussion, $h_a(\cdot)$ is trained with the standard loss function and regular random sampling strategy. In this section, we propose to use self-supervised learning to enhance the feature representation. Self-supervised learning is initially proposed for learning feature representations from a large-scale unlabelled dataset. Since the class information is unavailable, self-supervised learning usually relies on a pretext task as the surrogate training objective. Optimising towards a carefully chosen pretext objective can result in a representation that benefits downstream tasks. 

At first glance, it seems unnecessary to use self-supervised learning in a fully-supervised setting since we already have the ground-truth class labels. However, this view may be one-sided, which can be explained as: 
\textbf{1)} The tail classes alone do not have sufficient training samples to properly train the deep network and in our approach the  feature extractor part is (mainly) trained from samples in the head classes. In other words, our method is essentially built on the assumption that features (mainly) trained for head classes can also be useful for tail classes.
\textbf{2)} The primary goal of traditional supervised learning loss is to encourage a feature representation that supports a good separation of samples from different classes. Although we empirically observed that features trained for separating samples from class set A can generalise to class set B, it is unclear if using the supervised training towards A is optimal for achieving good performance on B or A+B. 
\textbf{3)} Self-supervised training is another existing feature learning method known to be able to learn feature presentations with cross-class generalisation capability, and it does not rely on the definition of visual classes. We postulate that it may be complementary to supervised training in feature representation learning, and we expect that using both training strategies will lead to better cross-class generalisation.

Motivated by the above consideration, we create another classifier $h_s(\cdot)$ attached to the shared feature extractor and employ a self-supervised training task as an additional auxiliary task. Specifically, we use the self-supervised method proposed in \cite{gidaris2018unsupervised}, where each input image $i$ is randomly rotated by one of four angles $\theta \in \{0^\circ, 90^\circ, 180^\circ, 270^\circ \}$ and the pretext task is to predict the rotation angle given the rotated input image. 

\subsection{Training \& Prediction}

As we address the dilemma of balancing head and tail classes by the auxiliary learning framework, we describe the full training and prediction details as follows: the ResNet model is employed as the feature extractor $\varphi(\cdot)$, which is shared among tasks, while fully-connected layers are utilised as classifiers $h(\cdot)$, $h_a(\cdot)$ and $h_s(\cdot)$ for each task.  The standard unweighted cross-entropy loss function is used in each task to compute the differences between predictions and targets. 

For each forward pass, two mini-batches are sampled by class-balanced and regular sampler respectively and then concatenated as one batch. The primary classification task accepts the class-balanced mini-batch $B_\text{CBS}$ as input to make predictions, while the random sampled one $B_\text{RRS}$ is sent to the regular random sampling auxiliary task. 
For self-supervised auxiliary task, the rotations are applied on the whole batch $B_\theta =[B_\text{CBS}; B_\text{RRS}]_\theta $. 
The final loss $L_{\text{Final}}$ of the deep network is computed as the weighted sum of the respective loss from each task:

\begin{equation}
    \begin{aligned}
        L_{\text{Final}} &=\lambda_1 l_{\text{CBS}}(h(\varphi(B_\text{CBS})), Y_\text{CBS})\\
        &+ \lambda_2 l_{\text{RRS}}(h_a(\varphi(B_\text{RRS})), Y_\text{RRS}) \\
        &+ \lambda_3 l_\text{SS}(h_s(\varphi(B_\theta)), Y_\theta)
    \end{aligned}
\end{equation}

We describe the full training algorithm of proposed model in Alg. \ref{CA_sample}. During the test, only the feature extractor $\varphi(\cdot)$ and the classifier $h(\cdot)$ from the primary task are retained for prediction. Therefore, compared to the standard classification network, no extra parameters are used in the test model.

\begin{algorithm}[t] 
    \centering
	\caption{Full Training Algorithm of Proposed \text{CBS}+\text{RRS}$^\dagger$+SS$^\dagger$.}
	\label{CA_sample} 
	\begin{algorithmic}[1]
		\Require
		\Statex $\{\mathcal{I}, \mathcal{Y}\}:$ Training Set, $| \mathcal{Y}|=C$; 
		\Statex $M:$ Total iterations; $S:$ Input mini-batch size;
		\Statex $Q_C:$ Ordered class list, $|Q_C|=C$;
		\Statex $\{Q_y\}:$ Ordered sample list per class, $\forall y\in \mathcal{Y}$;
		\Statex $Z:$ The number of images per class to be sampled in each mini-batch by CBS;
		\Statex $\varphi(\cdot):$ Feature extractor; 
		\Statex $h(\cdot), h_a(\cdot), h_s(\cdot):$ Classifiers;
		\Statex
		
		\For {$m\in [1, M]$}
    		\State Randomly sample $k = S/Z$ classes $\mathcal{Y}_k$ from the $Q_C$ without replacement;
    		\State $B_{\text{CBS}} = \{ \}$;
    		\For {$y\in \mathcal{Y}_k$}
    		    \State Randomly sample $Z$ samples $\{I_Z\}$ from $Q_{y}$ without replacement; 
    		    \State $B_\text{CBS}$=\{$B_\text{CBS}$; $\{I_Z\}$\};
    		\EndFor
    		\State Randomly sample $S$ images as $B_{\text{RRS}}$;
    		\State $B_\theta$=Rotate$([B_{\text{CBS}}; B_{\text{RRS}}])$;
    		\State $l_{\text{CBS}} = l(h(\varphi(B_\text{CBS})), Y_\text{CBS})$; 
    		\quad $l_{\text{RRS}} = l(h_a(\varphi(B_\text{RRS})), Y_\text{RRS})$; 
    		\State $l_{\text{SS}} = l(h_s(\varphi(B_\theta), Y_\theta)$; 
    		\quad $L_{\text{Final}} = \lambda_1 l_{\text{CBS}} + \lambda_2 l_{\text{RRS}} + \lambda_3 l_{\text{SS}}$;
    		\State Optimise $L_{\text{Final}}$ by SGD.
		\EndFor
	\end{algorithmic}
\end{algorithm}

\section{Experiment}
\label{exp}
We present experimental results in this section and investigate the effectiveness of the proposed model through extensive ablation studies. 

\subsection{Datasets}
The proposed model is evaluated on the long-tailed version (-LT) of two benchmark datasets: ImageNet-LT and Places-LT \cite{liu2019large} with various backbone feature extractors. Both LT datasets are constructed by sampling from original datasets (ImageNet-2012 \cite{deng2009imagenet} and Places-2 \cite{zhou2016places}) under the Pareto distribution with $\alpha$=6. The ImageNet-LT contains $N$=185,846 images from $C$=1000 classes, among which 115,846/ 20,000/ 50,000 images are used for training/ validation/ test. The number of images per class $N_c$ ranges from minimal 5 to maximum 1280. For Places-LT, there are 106,300 images from 365 categories, with training, validation and test splits of 62,500/ 7,300/ 36,500 images. The imbalance is more severe than ImageNet-LT with $N_c$ ranges from 5 to 4980. For both datasets, the test sets are made balanced. 

\subsection{Evaluation Metrics}
Following the evaluation protocols in \cite{liu2019large}, both overall top-1 classification accuracy and shot-wise accuracy are computed. The overall accuracy is computed over all classes, while the test set is split into three sub-sets for evaluating shot-wise accuracy, namely many shot (classes with $N_c\geq 100$), low shot ($N_c \leq 20$) and medium shot ($N_c\in (20, 100)$). The shot-wise accuracy aims at monitoring the behaviours of the proposed model on the different portions of the imbalanced distribution. For clear demonstrations, shot-wise splits of ImageNet-LT is shown in the distribution figure in Fig. \ref{4_ImageNet_gain_chart}.

\subsection{Compared Methods}
We compare our models with several state-of-the-art methods, as well as two straightforward baselines:

\textbf{RRS-Only}: This baseline is equivalent to the conventional supervised training for image classification tasks, with regular randomly sampled batches as inputs.

\textbf{CBS-Only}: This baseline is only to apply class-balanced sampling strategy to train the whole network without any auxiliary task.

\textbf{Ours}: CBS+$\text{RRS}^{\dagger}$ \footnote{$\dagger$ indicates the task is an auxiliary one.} is the basic version of our method, which uses the regular random sampling branch to construct auxiliary learning task. 
CBS+$\text{RRS}^{\dagger}$+$\text{SS}^{\dagger}$ is our extended method with both regular random sampling and self-supervised learning as auxiliary tasks.

\textbf{SOTA}: Various state-of-the-art methods belonging to different categories are used for comparisons. In particular, we compare against methods based on metric learning, including lifted structure loss \cite{DBLP:conf/cvpr/SongXJS16}, triplet Loss \cite{DBLP:journals/corr/HofferA14} and proxy static loss \cite{movshovitz2017no}; methods based on hard example mining and feature regularisation, including focal loss \cite{DBLP:conf/iccv/LinGGHD17}, range loss \cite{DBLP:conf/iccv/ZhangFWLQ17}, few shot learning without forgetting (FSLwF) \cite{DBLP:conf/cvpr/GidarisK18} and memory enhancement (MemoryNet) \cite{liu2019large}; as well as cost-sensitive loss \cite{haixiang2017learning}. For fair comparisons, we adopt the same feature extractor $\varphi(\cdot)$ as \cite{liu2019large} in all compared models, that is, ResNet as feature backbones and followed with an average-pooling layer and a 512-dim Fc layer. 

\subsection{Experimental Results}
Four groups of experiments are conducted. To make a fair comparison with results reported in \cite{liu2019large}, the ResNet-10 trained from scratch and the pre-trained ResNet-152 are used for ImageNet-LT and Places-LT respectively. Moreover, to fully investigate the effectiveness of the proposed method with different backbones, we also train ResNet-50 models from scratch on both ImageNet-LT and Places-LT datasets.
We set each sampled class has $Z=4$ images in class-balanced sampling with $\lambda_1:$$\lambda_2:$$\lambda_3$= 0.5:1:1 in our extended model. SGD with learning rate decays by following the cosine learning schedule is used for optimisation. Learning rates, number of epochs and batches sizes are set by referring to \cite{liu2019large} and cross-validated on the validation sets. The experimental results are shown in Tab. \ref{report_res} and \ref{own_res}. 

\begin{table}[t]
\centering
\resizebox{\linewidth}{!}{
\begin{tabular}{c|cccc|cccc}
\Xhline{2\arrayrulewidth}
Dataset: Backbone & \multicolumn{4}{c}{ImageNet-LT: ResNet-10} & \multicolumn{4}{c}{Places-LT: ResNet-152} \\
Top-1 Accuracy (\%) & $>=100$ & $<100\& >20$ & $<=20$ &  & $>=100$ & $<100\& >20$ & $<=20$ &  \\
Methods & Many shot & Medium shot & Low shot & \textbf{Overall} & Many shot & Medium shot & Low shot & \textbf{Overall} \\\hline\hline
Lifted Loss \cite{DBLP:conf/cvpr/SongXJS16} & 35.8 & 30.4 & 17.9 & 30.8 & 41.1 & 35.4 & 24.0 & 35.2 \\
Focal Loss \cite{DBLP:conf/iccv/LinGGHD17} & 36.4 & 29.9 & 16.0 & 30.5 & 41.1 & 34.8 & 22.4 & 34.6 \\
Range Loss \cite{DBLP:conf/iccv/ZhangFWLQ17} & 35.8 & 30.3 & 17.6 & 30.7 & 41.1 & 35.4 & 23.2 & 35.1 \\
Cost-Sensitive \cite{haixiang2017learning} & 37.2 & 30.9 & 15.9 & 31.2 & 38.8 & 36.9 & 20.9 & 34.2 \\
FSLwF \cite{DBLP:conf/cvpr/GidarisK18} & 40.9 & 22.1 & 15.0 & 28.4 & 43.9 & 29.9 & 29.5 & 34.9  \\
MemoryNet \cite{liu2019large} & 43.2 & 35.1 & 18.5 & 35.6 & 44.7 & 37.0 & 25.3 & 35.9 \\
MemoryNet$^*$ \cite{liu2019large} & 47.9 & 37.0 & 16.6 & 38.3 & 44.4 & 39.3 & 27.3 & 38.6 \\
\hline\hline
RRS-Only & \textbf{\textcolor{red}{53.4}} & 23.5 & 4.3 & 32.4 & \textbf{\textcolor{red}{47.4}} & 27.1 & 10.5 & 31.0 \\
CBS-Only & 35.4 & 31.5 & \textbf{\textcolor{blue}{18.6}} & 31.2 & \textbf{\textcolor{blue}{46.9}} & 34.9 & 18.1 & 35.7 \\
CBS+RRS$\dagger$ & 51.0 & \textbf{\textcolor{blue}{38.5}} & \textbf{\textcolor{red}{21.0}} & \textbf{\textcolor{blue}{40.9}} & 42.5 & \textbf{\textcolor{blue}{40.3}} & \textbf{\textcolor{blue}{32.2}} & \textbf{\textcolor{blue}{39.4}} \\
CBS+RRS$^\dagger$+SS$^\dagger$ & \textbf{\textcolor{blue}{52.4}} & \textbf{\textcolor{red}{39.5}} & \textbf{\textcolor{red}{21.0}} & \textbf{\textcolor{red}{41.9}} & 42.5 & \textbf{\textcolor{red}{40.6}} & \textbf{\textcolor{red}{32.6}} & \textbf{\textcolor{red}{39.6}} \\
\Xhline{2\arrayrulewidth}
\end{tabular}}
\vspace{5pt}
\caption{Experimental results on ImageNet-LT and Places-LT datasets with backbone ResNet-10 trained from scratch and pre-trained ResNet-152. The best and second-best scores are noted in red and blue, respectively.}
\label{report_res}
\end{table}

\begin{table}[t!]
\centering
\resizebox{\linewidth}{!}{
\begin{tabular}{c|cccc|cccc}
\Xhline{2\arrayrulewidth}
Dataset: Backbone & \multicolumn{4}{c}{ImageNet-LT: ResNet-50} & \multicolumn{4}{c}{Places-LT: ResNet-50} \\
Top-1 Accuracy (\%) & $>=100$ & $<100\& >20$ & $<=20$ &  & $>=100$ & $<100\& >20$ & $<=20$ &  \\
Methods & Many shot & Medium shot & Low shot & \textbf{Overall} & Many shot & Medium shot & Low shot & \textbf{Overall} \\\hline\hline
Triplet Loss \cite{DBLP:journals/corr/HofferA14} & 48.3 & 39.5 & 19.3 & 40.0 & 30.4 & 27.8 & 6.8 & 24.3 \\
Focal Loss \cite{DBLP:conf/iccv/LinGGHD17} & 49.0 & 39.2 & 19.1 & 40.1 & 26.2 & 22.8 & 6.2 & 20.5 \\
Proxy Loss \cite{movshovitz2017no} & 49.0 & 38.9 & 17.7 & 39.8 & 30.3 & 24.4 & 4.8 & 22.4 \\
Cost-Sensitive \cite{haixiang2017learning} & 48.0 & 38.8 & 19.6 & 39.6 & 27.1 & 26.4 & \textbf{\textcolor{red}{16.9}} & 24.6\\
MemoryNet$^*$ \cite{liu2019large} & 56.6 & 39.3 & 16.4 & 42.7 & 34.1 & 26.2 & 8.2 & 25.3\\
\hline\hline
RRS-Only & \textbf{\textcolor{red}{61.3}} & 32.2 & 7.2 & 39.9 & \textbf{\textcolor{red}{39.1}} & 12.0 & 0.3 & 19.3\\
CBS-Only & 51.1 & 38.2 & 16.7 & 40.1 & 32.1 & 22.9 & 4.6 & 22.4\\
CBS+RRS$^\dagger$ & 59.0 & \textbf{\textcolor{blue}{43.5}} & \textbf{\textcolor{blue}{22.8}} & \textbf{\textcolor{blue}{46.5}} & 35.6 & \textbf{\textcolor{blue}{29.3}} & 11.3 & \textbf{\textcolor{blue}{27.8}}\\
CBS+RRS$^\dagger$+SS$^\dagger$ & \textbf{\textcolor{blue}{60.2}} & \textbf{\textcolor{red}{44.0}} & \textbf{\textcolor{red}{23.1}} & \textbf{\textcolor{red}{47.3}} & \textbf{\textcolor{blue}{35.8}} & \textbf{\textcolor{red}{31.0}} & \textbf{\textcolor{blue}{12.7}} & \textbf{\textcolor{red}{28.9}}\\
\Xhline{2\arrayrulewidth}
\end{tabular}}
\vspace{5pt}
\caption{Experimental results on ImageNet-LT and Places-LT dataset with backbone ResNet-50 trained from scratch. The best and second-best scores are noted in red and blue, respectively.}
\label{own_res}
\end{table}
First, we discuss the comparisons against baselines on both datasets in Tab. \ref{report_res} and \ref{own_res}. As expected, the standard supervised training RRS-Only achieves high accuracy on the many shot split, yet extremely low performances on other two splits. Using CBS-only, on the other extreme, achieves reasonable performance on medium shot and low shot splits, but performs poorly on many shot. This result validates that directly applying CBS makes the head-classes under-represented. As seen, our CBS+$\text{RRS}^{\dagger}$ method significantly improves the accuracy of medium and low shot splits compared to the baseline RRS-Only. Compared to CBS-Only, it effectively prevents the under-representation of head classes, results in superior performances on the many shot split. 

By further comparing CBS+$\text{RRS}^{\dagger}$ against other existing methods, we find the proposed method achieves the overall best performance. For example, we outperform the second best method, MemoryNet \cite{liu2019large} by averagely 3\%, despite the latter uses more complicated algorithm designs and model parameters. Also, it is observed that our models excel at each split. This clearly shows that the proposed method is excellent in balancing the head and tail classes. For fair comparisons, in addition to its reported results, we also reproduce MemoryNet \cite{liu2019large} results (indicated by $^*$) using authors' code \footnote{https://github.com/zhmiao/OpenLongTailRecognition-OLTR} with our training strategy. Furthermore, from Tab. \ref{report_res} and \ref{own_res} we can also observe that, in general, adding self-supervised auxiliary task SS$^\dagger$ can bring additional improvements.

\begin{figure}[t]
    \vspace{10pt}
	\centering
	\resizebox{1.0\linewidth}{!}{
		\begin{tabular}{cc}
			\includegraphics[width=0.5\linewidth]{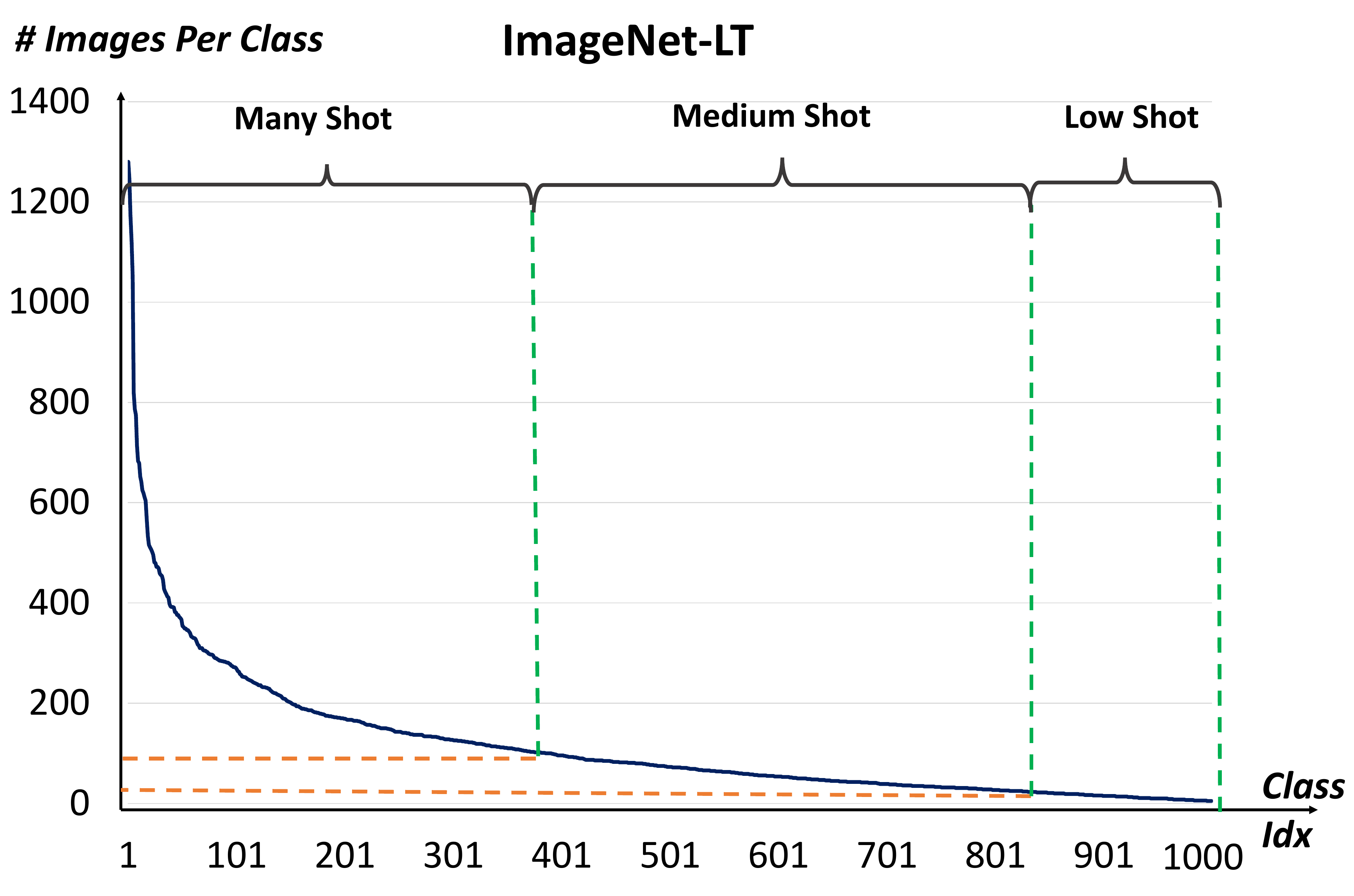} & 
			\includegraphics[width=0.5\linewidth]{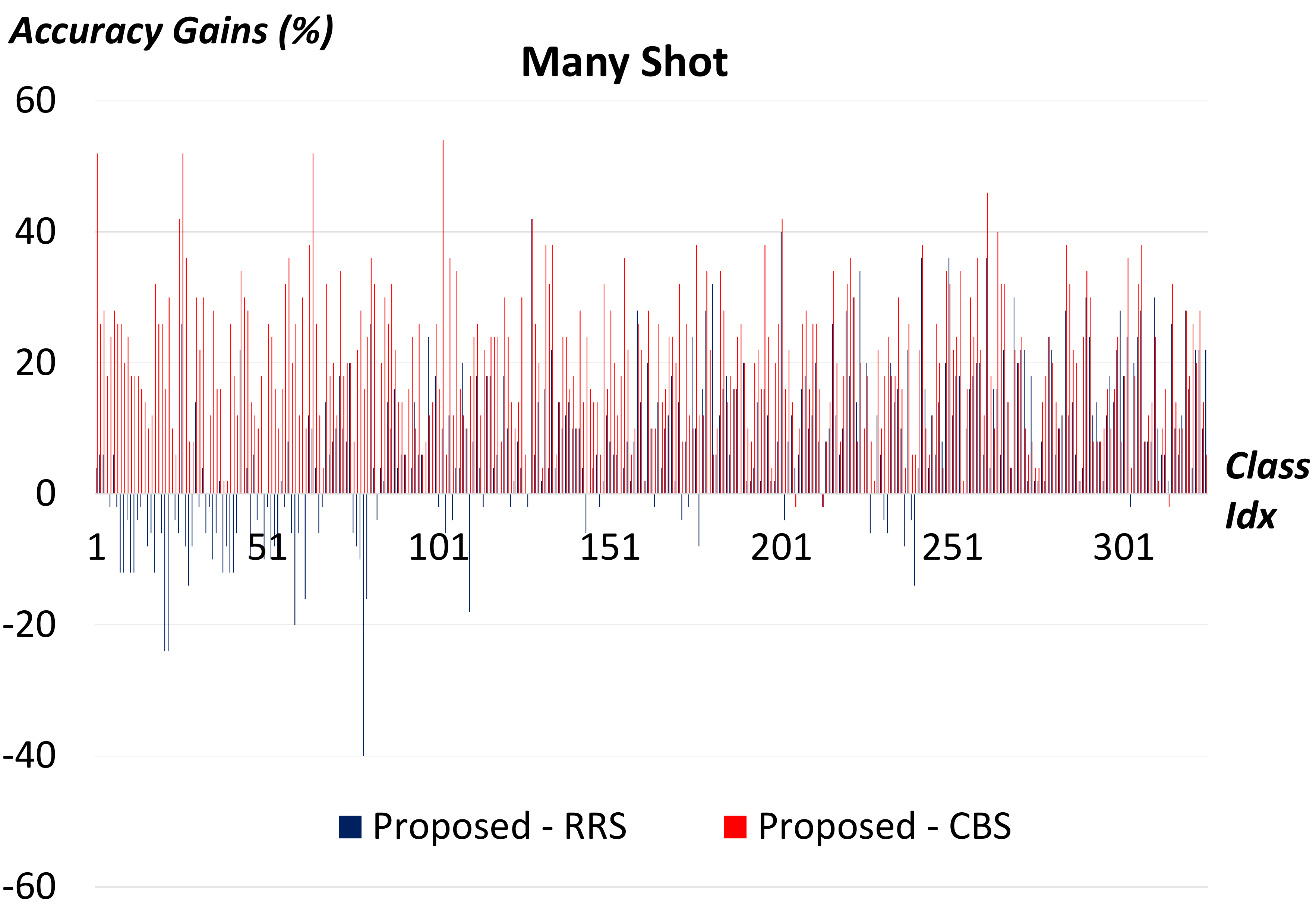}\\\\
			\includegraphics[width=0.5\linewidth]{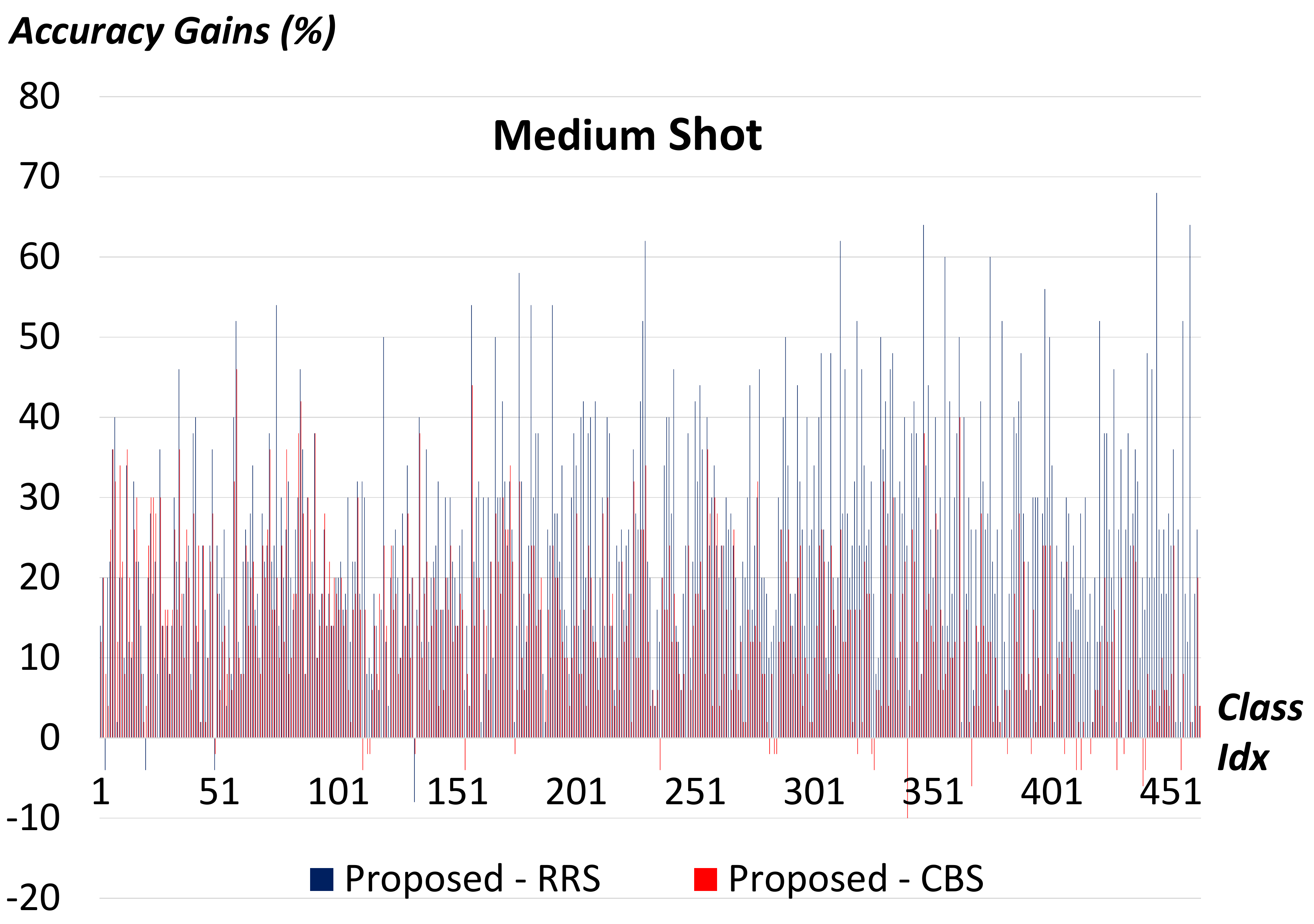} & 
			\includegraphics[width=0.5\linewidth]{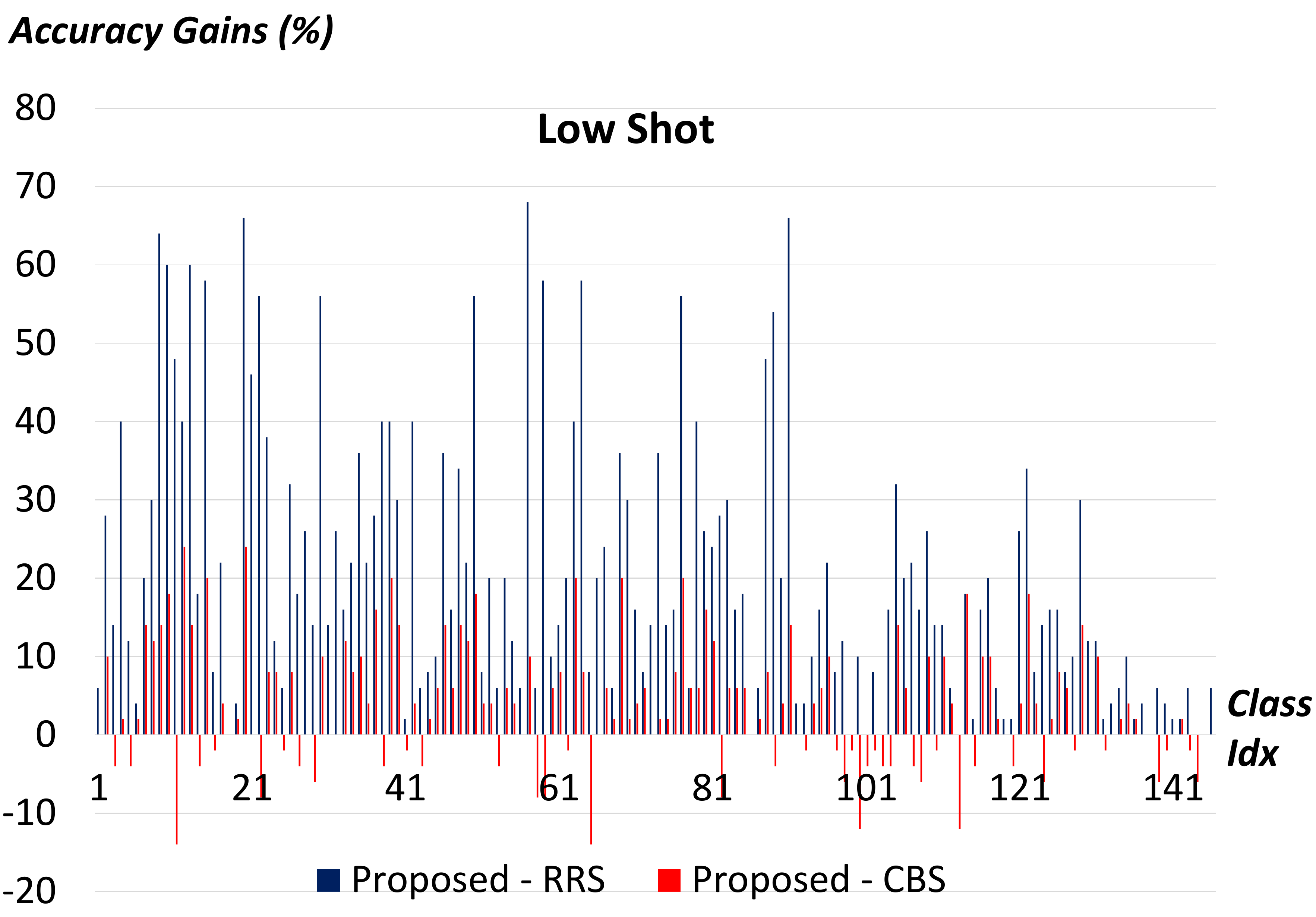}
			
	\end{tabular}}
	\caption{Accuracy gains of the proposed model comparing to the baselines RRS-Only and CBS-Only over all classes on ImageNet-LT, which are represented by blue bars and red bars respectively. The classes are sorted \textit{w.r.t.} frequencies. The upper left figure shows the long-tailed dataset distribution.
	}
	\label{4_ImageNet_gain_chart}
\end{figure}

To demonstrate the performance gains of proposed CBS+$\text{RRS}^{\dagger}$ over two important baselines RRS-Only and CBS-Only in each class, we compute the class-wise accuracy gain, which is defined as: ACC$_\text{proposed} $-ACC$_\text{baseline}$, where ACC$_\text{proposed}$ and $_\text{baseline}$ are the prediction accuracy of the proposed method and the to-be-compared baselines respectively. 
The result of ImageNet-LT is shown in Fig. \ref{4_ImageNet_gain_chart}. The left y-axis indicates the accuracy gains and the right axis is the class frequencies. The bar charts correspond to the comparisons against RRS-Only (blue) and CBS-Only (red) the class indexes are sorted \textit{w.r.t} their frequencies. We separate all classes into three splits to be corresponded with many, medium and low shot respectively.
It is clear that CBS+$\text{RRS}^{\dagger}$ leads to significant improved performances for most classes, especially for head classes (many shot) when comparing against CBS-Only, and tail classes (medium and low shots) when comparing against RRS-Only.

\begin{table*}[t!]
\centering
\resizebox{\linewidth}{!}{
\begin{tabular}{c|cccc|cccc}
\Xhline{2\arrayrulewidth}
Dataset: Backbone & \multicolumn{4}{c}{ImageNet-LT: ResNet-10} & \multicolumn{4}{c}{Places-LT: ResNet-50} \\
Top-1 Accuracy (\%) & $>=100$ & $<100\& >20$ & $<=20$ &  & $>=100$ & $<100\& >20$ & $<=20$ &  \\
Methods & Many shot & Medium shot & Low shot & \textbf{Overall} & Many shot & Medium shot & Low shot & \textbf{Overall} \\\hline\hline
RRS-Only  & \textbf{53.4} & 23.5 & 4.3 & 32.4 & \textbf{39.1} & 12.0 & 0.3 & 19.3 \\
CBS-Only & 35.4 & 31.5 & 18.6 & 31.2 & 32.1 & 22.9 & 4.6 & 22.4 \\
ftRRS+CBS & 48.3 & 35.2 & 16.4 & 37.6 & 29.3 & 27.6 & 10.3 & 24.6 \\
CBS+$\text{RRS}^{\dagger}$ & 51.0 & \textbf{38.5} & \textbf{21.0} & \textbf{40.9} & 35.6 & \textbf{29.3} & \textbf{11.3} & \textbf{27.8} \\\Xhline{2\arrayrulewidth}
\end{tabular}}
\vspace{5pt}
\caption{Experimental results of auxiliary learning vs stage-wise training on two datasets. -Only models are trained from scratch, while the feature extractor of ftRRS+CBS is initialised from RRS-Only weights.}
\label{smp_ft}
\end{table*}

\begin{table*}[t!]
\centering
\resizebox{\linewidth}{!}{
\begin{tabular}{c|cccc|cccc}
\Xhline{2\arrayrulewidth}
Dataset: Backbone & \multicolumn{4}{c}{ImageNet-LT: ResNet-10} & \multicolumn{4}{c}{Places-LT: ResNet-50} \\
Top-1 Accuracy (\%) & $>=100$ & $<100\& >20$ & $<=20$ &  & $>=100$ & $<100\& >20$ & $<=20$ &  \\
CBS:RRS Weights & Many shot & Medium shot & Low shot & \textbf{Overall} & Many shot & Medium shot & Low shot & \textbf{Overall} \\\hline\hline
0:1 (RRS-Only) & \textbf{53.4} & 23.5 & 4.3 & 32.4 & \textbf{39.1} & 12.0 & 0.3 & 19.3 \\
0.5:1 & 51.0 & \textbf{38.5} & \textbf{21.0} & \textbf{40.9} & 35.6 & \textbf{29.3} & \textbf{11.3} & \textbf{27.8} \\
1:1 & 49.4 & 37.7 & 19.8 & 39.6 & 37.8 & 27.6 & 7.8 & 27.1 \\
1:0.5 & 46.0 & 35.5 & 17.8 & 37.0 & 34.9 & 26.9 & 6.9 & 25.6 \\
1:0 (CBS-Only) & 35.4 & 31.5 & 18.6 & 31.2 & 32.1 & 22.9 & 4.6 & 22.4 \\\Xhline{2\arrayrulewidth}
\end{tabular}}
\vspace{5pt}
\caption{Experimental results of ablation studies of different CBS:RRS ratios on two datasets. All models are trained under CBS+RRS$^\dagger$ framework.}
\label{weight_ratio}
\end{table*}

\begin{table}[t!]
\centering
\resizebox{\linewidth}{!}{
\begin{tabular}{c|cccc|cccc}
\Xhline{2\arrayrulewidth}
Dataset: Backbone & \multicolumn{4}{c}{ImageNet-LT: ResNet-10} & \multicolumn{4}{c}{Places-LT: ResNet-50} \\
Top-1 Accuracy (\%) & $>=100$ & $<100\& >20$ & $<=20$ &  & $>=100$ & $<100\& >20$ & $<=20$ &  \\
Methods & Many shot & Medium shot & Low shot & \textbf{Overall} & Many shot & Medium shot & Low shot & \textbf{Overall} \\\hline\hline
w/o Rotation & 51.0 & 38.5 & 21.0 & 40.9 & 35.6 & 29.3 & 11.3 & 27.8\\
Rotation Augmentation & 50.5 & 39.1 & \textbf{22.1} & 41.0 & 35.3 & 28.2 & 11.4 & 27.2 \\
CBS+$\text{RRS}^{\dagger}$+SS$^\dagger$ & \textbf{52.4} & \textbf{39.5} & 21.0 & \textbf{41.9} & \textbf{35.8} & \textbf{31.0} & \textbf{12.7} & \textbf{28.9}\\\Xhline{2\arrayrulewidth}
\end{tabular}}
\vspace{5pt}
\caption{Experimental results of using rotations as the pure data augmentation and self-supervised auxiliary task.}
\label{SS}
\end{table}

\subsection{Ablation Study}
In this section, we conduct extensive ablation studies to investigate the effectiveness of the proposed auxiliary tasks. To observe results free from the influence of pre-trained weights from additional datasets, all models are trained from scratch unless otherwise stated.

\textbf{Auxiliary Learning vs Stage-Wise Training. }
We first investigate the necessity of joint training of the primary task and auxiliary task. To verify this, we compare the proposed method against stage-wise training, which firstly trains the network with regular random sampling strategy and then fine-tunes with the CBS strategy. We denote this baseline as ftRRS+CBS and make comparisons in Tab. \ref{smp_ft}.
As seen, this stage-wise training strategy can achieve improved accuracy on medium and low shot splits compared to RRS-Only. This improvement can be largely attributed to the CBS strategy adopted at the fine-tuning stage. Also, compared to CBS-Only, ftRRS+CBS achieves a better result on the many shot split. This may be because the head class information has already been incorporated during the first stage training. However, when compared with CBS+$\text{RRS}^{\dagger}$, the stage-wise strategy still achieves the inferior result, which clearly shows the benefit of joint training. 

\textbf{Sampling Strategy $\lambda_1$:$\lambda_2$ Weight Ratio. }
Our method needs to set weights for different loss terms. In this section, we also explore the impact of different CBS:RRS weight ratios $\lambda_1$:$\lambda_2$ in the CBS+RRS$^\dagger$ model. We test different ratios ranging from 0:1 to 1:0, where 0:1 is equivalent to the RRS-Only baseline, and 1:0 is equivalent to the CBS-Only baseline. As observed in Tab. \ref{weight_ratio}, in general, the model can achieve promising results as long as two tasks are jointly trained. The highest overall performance strikes when $\lambda_1$:$\lambda_2$ = 0.5:1, which means that we should let RRS plays a major role in training the feature representation. This observation supports our motivation of introducing the RRS auxiliary branch. 

\textbf{Self-Supervised Learning or Data Augmentation Strategy? }
Given its simplicity and effectiveness, rotation-based self-supervised learning \cite{gidaris2018unsupervised} is used as the auxiliary task to enhance feature learning. One may suspect that the benefit of introducing this task is essentially a data augmentation strategy. To investigate this, we report the performance of directly using rotations for data augmentation. Specifically, we feed the four versions of rotated images into the CBS+$\text{RRS}^{\dagger}$ model to ensure the same inputs as in CBS+$\text{RRS}^{\dagger}$+$\text{SS}^{\dagger}$. As Tab. \ref{SS} shows, rotation-based data augmentation cannot bring significant improvements over the baseline. This validates that the benefit of using rotation-based self-supervised learning cannot be simply explained by the rotation augmentation.

\subsection{Visualisation}
To further demonstrate the effectiveness of the proposed approach of employing different sampling strategies for feature and classifier training, we examine the qualities of the learned features and classifiers separately. We take models trained on ImageNet-LT as an example and visualise the learned features and classifier weights.

\begin{figure}[t]
	\centering
	\resizebox{0.88\linewidth}{!}{
		\begin{tabular}{cc}
		\centering
			\includegraphics[width=0.4\linewidth, height=0.4\linewidth]{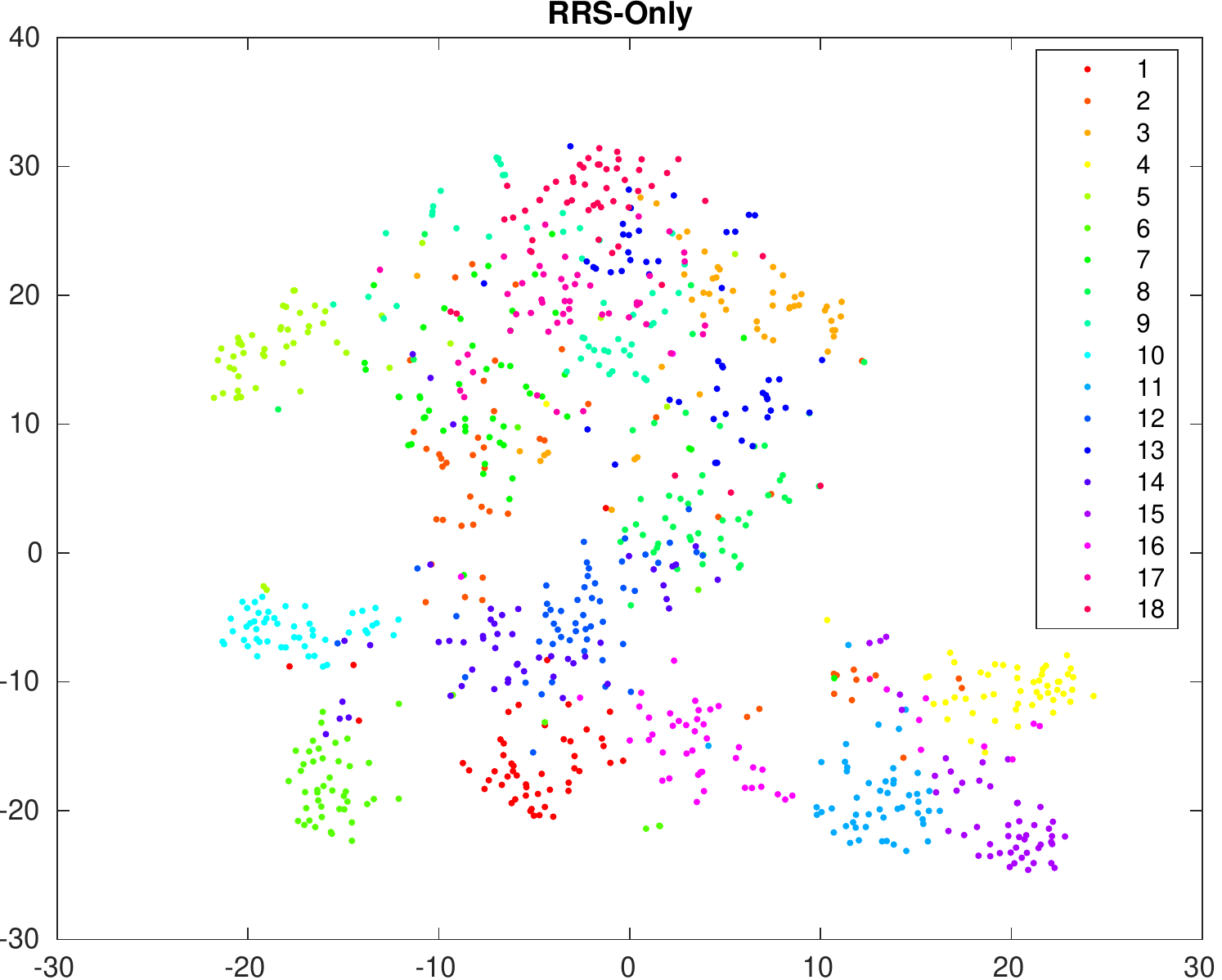} & 
			\includegraphics[width=0.4\linewidth, height=0.4\linewidth]{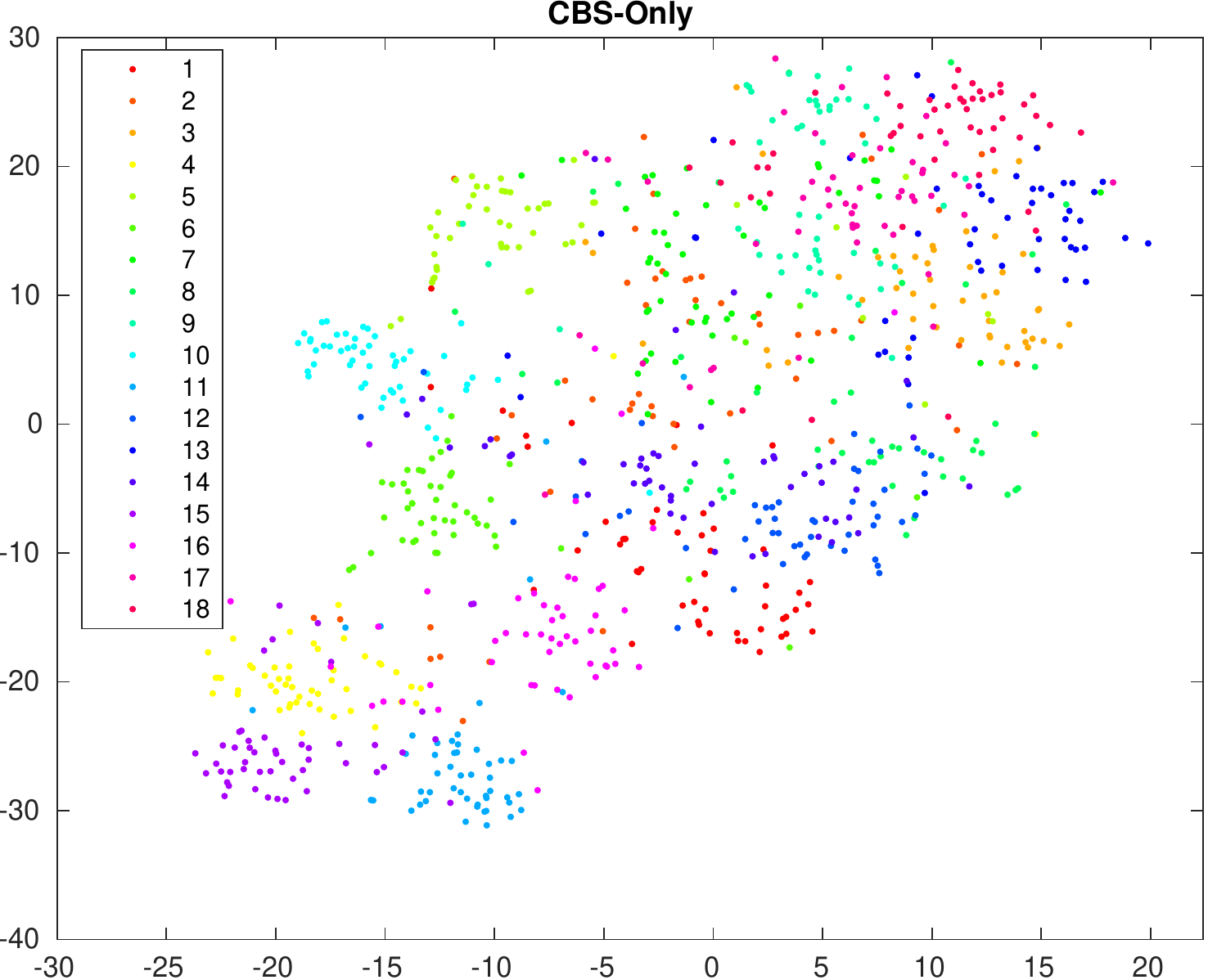}\\\\
			\includegraphics[width=0.4\linewidth, height=0.4\linewidth]{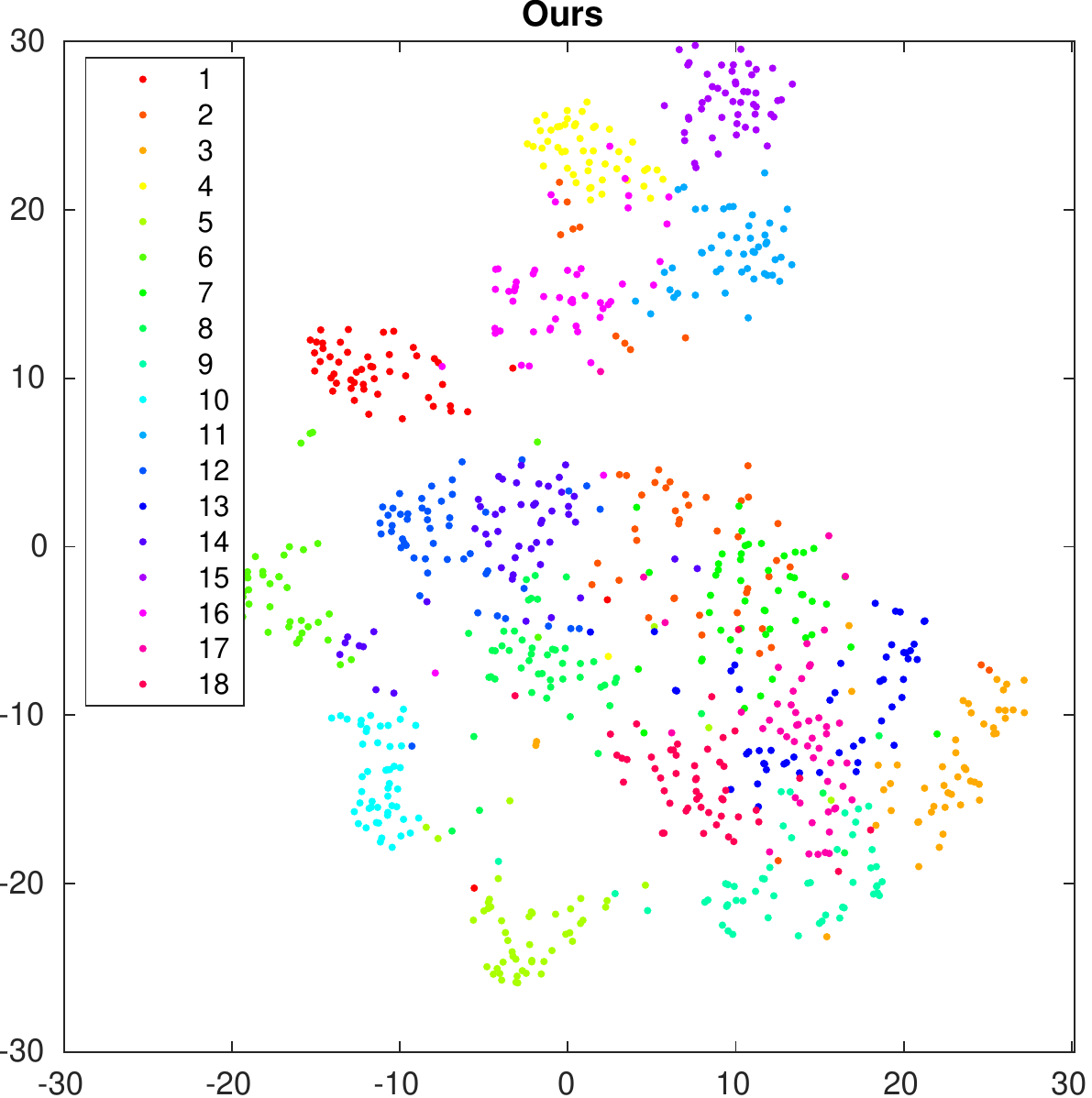} & 
			\includegraphics[width=0.41\linewidth]{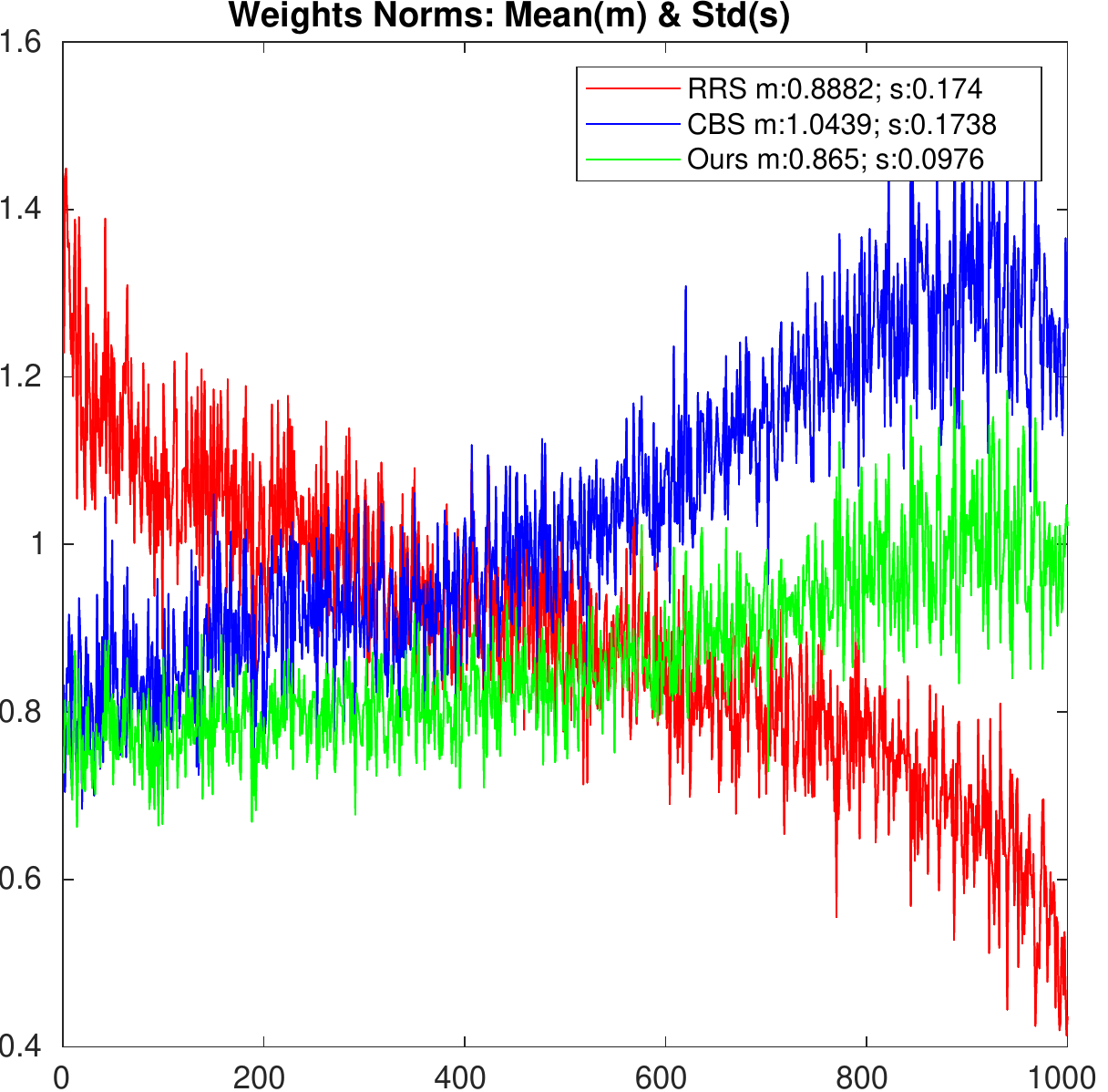}
	\end{tabular}}
	\caption{The visualisation of learned features and classifiers. In the first three sub-figures, 18 classes are evenly selected from three splits for feature visualisation by performing t-SNE. The forth sub-figure shows the distribution of the classifier weights from three compared models, classes are sorted \textit{w.r.t.} their frequencies. 
	}
	\label{6_visualise}
\end{figure}

\textbf{Feature. }
One way to visually examine the quality of the image feature is to perform visualisation by t-Distributed Stochastic Neighbor Embedding (t-SNE) \cite{maaten2008visualizing}. That is, we select test image features from 18 classes and visualise them in Fig. \ref{6_visualise}, among which, six classes are selected from each shot (many, medium and low, from top to bottom), and 50 images are used in each class.
As expected, compared to both baselines, our model takes advantage of both sampling strategies and shows the strongest feature learning ability as features are well separated for the majority of the classes from all shots.

\textbf{Classifier Weights. }
The $l_2$-norm of the classifier weight has been used for identifying the biases among classifiers in previous works \cite{salimans2016weight,guo2017one,smirnov2017doppelganger,wu2017low}, \textit{i.e.} the ideal multi-class classifiers should have evenly distributed weight norms. 
Therefore, we compute the $l_2$-norms of the 1000-way classifier and plot them in Fig. \ref{6_visualise} with respect to their class frequencies. As we can see, the distribution of RRS-Only norms has the similar pattern as the long-tailed dataset distribution, while CBS-Only has a smaller bandwidth and reverse distribution, as the tail classes are over-sampled to achieve class balance. Our proposed model shows the most balanced distribution with the smallest variation among different classes since we take advantage of both sampling strategies.

\section{Conclusion}
In this paper, we address the long-tailed recognition by fitting it into a simple-yet-effective auxiliary learning framework. The dilemma of balancing the head and tail classes training is analysed, and the class-balanced sampling strategy is adopted in the primary task to tackle the unfair training of tail classes, while the regular random sampling is used in the auxiliary task to jointly prevent the ill-fitting during feature learning. To further enhance the feature representation, self-supervised learning is explored as an additional auxiliary task. Comparisons against state-of-the-art methods and extensive ablation studies verify the effectiveness of proposed models.

% \clearpage
\bibliographystyle{splncs04}
\bibliography{egbib}
\end{document}